\documentclass[11pt]{article}

\usepackage{amsmath}
\usepackage{amssymb}
\usepackage{multirow}

\usepackage{makecell}

\usepackage{array}
\usepackage[final ]{acl}

\usepackage{tabularx}
\usepackage{times}
\usepackage{latexsym}
\usepackage{enumitem}
\usepackage{booktabs}
\usepackage{seqsplit}
\usepackage[T1]{fontenc}

\usepackage[utf8]{inputenc}

\usepackage{microtype}
\usepackage{seqsplit}  
\usepackage{inconsolata}

\usepackage{graphicx}
\usepackage{xcolor}  
\usepackage{tikz}

\usetikzlibrary{arrows.meta, positioning, fit, backgrounds, shapes.geometric, calc}
 
\definecolor{gptblue}{RGB}{215,232,250}
\definecolor{gptblueline}{RGB}{47,110,180}
\definecolor{rulegray}{RGB}{232,232,232}
\definecolor{rulegrayline}{RGB}{90,90,90}
\definecolor{humanorange}{RGB}{253,229,203}
\definecolor{humanorangeline}{RGB}{204,120,20}
\definecolor{fargreen}{RGB}{222,241,222}
\definecolor{fargreenline}{RGB}{56,130,56}
\definecolor{fppyellow}{RGB}{250,240,200}
\definecolor{fppyellowline}{RGB}{176,141,17}
\definecolor{ftrpurple}{RGB}{233,222,245}
\definecolor{ftrpurpleline}{RGB}{110,60,160}
\definecolor{dpored}{RGB}{248,222,222}
\definecolor{dporedline}{RGB}{176,50,50}
 
\newcommand{\circled}[1]{\tikz[baseline=(char.base)]{\node[draw, circle, inner sep=0.4pt, minimum size=1.7ex, font=\scriptsize] (char) {#1};}}

\title{Mind the Gap: Theory-of-Mind-Grounded Friction for Epistemic Alignment}

\author{
  Yifan Zhu
  \and
  Kyeongmin Rim
  \and
  James Pustejovsky \\
  Brandeis University, Waltham, MA, USA \\
  \texttt{\{zhuyifan, krim, jamesp\}@brandeis.edu}
}

\begin{document}
\maketitle

\begin{abstract}


Productive dialogue alignment requires distinguishing \emph{surface coordination} (acknowledgments and smooth task progression) from \emph{epistemic alignment} (convergence of belief states); standard preference-based methods typically optimize response-level preferences without explicitly modeling the latter. We operationalize Theory-of-Mind (ToM) inference as a control signal within Frictive Policy Optimization by extracting, at each referring expression, a four-part belief structure: the speaker's intended referent, the addressee's interpretation, and each participant's model of the other's belief. This makes friction mechanically computable from epistemic-state comparisons, capturing \emph{silent divergence}, where both participants proceed confidently while grounding to different referents. We evaluate the signal at two levels. At the representation level, ablating the second-order channel reduces misunderstanding recall from $65\%$ to $26\%$. At the policy level, reward-shaping (FAR) and trust-region (FTR) variants improve intervention F1 and warranted-context calibration over DPO, with Brier scores independently supporting the calibration gains. Across three training runs, FAR and FTR remain substantially more stable, whereas DPO varies widely and can degrade intervention competence already present in the base policy. Thus, ToM-grounded friction provides a trainable signal for context-sensitive intervention under referential belief divergence.

\end{abstract}

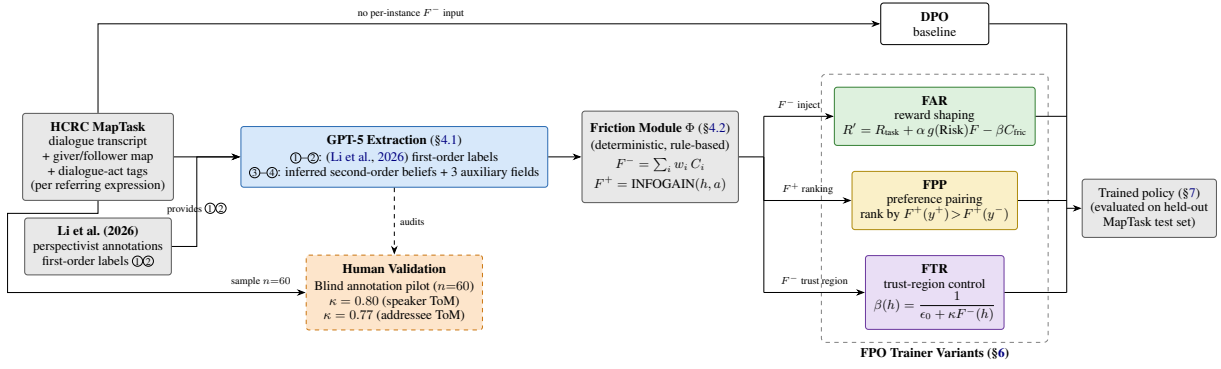
\begin{figure*}[!tp]
\centering
\resizebox{\textwidth}{!}{%
\begin{tikzpicture}[
    font=\small,
    node distance=8mm and 12mm,
    >={Stealth[length=2.5mm]},
    every node/.style={align=center},
    box/.style={
        rectangle, rounded corners=2pt, draw, minimum width=32mm,
        minimum height=10mm, inner sep=2mm, line width=0.7pt
    },
    inputbox/.style={box, fill=rulegray, draw=rulegrayline},
    gptbox/.style={box, fill=gptblue, draw=gptblueline},
    rulebox/.style={box, fill=rulegray, draw=rulegrayline},
    humanbox/.style={box, fill=humanorange, draw=humanorangeline, dashed},
    farbox/.style={box, fill=fargreen, draw=fargreenline, minimum width=26mm},
    fppbox/.style={box, fill=fppyellow, draw=fppyellowline, minimum width=26mm},
    ftrbox/.style={box, fill=ftrpurple, draw=ftrpurpleline, minimum width=26mm},
    dpobox/.style={box, minimum width=26mm},
    groupbox/.style={
        rectangle, rounded corners=3pt, draw=black!50, dashed,
        inner sep=3mm, label={[font=\bfseries\footnotesize]below:#1}
    },
    arr/.style={-{Stealth}, line width=0.7pt},
    legenditem/.style={rectangle, minimum width=4mm, minimum height=3mm, draw, line width=0.5pt}
]
 
\node[inputbox] (mapinput)
    {\textbf{HCRC MapTask}\\
     \footnotesize dialogue transcript\\
     \footnotesize + giver/follower map\\
     \footnotesize + dialogue-act tags\\
     \footnotesize (per referring expression)};
\node[inputbox, below=4mm of mapinput] (liinput)
    {\textbf{Li et al.\ (2026)}\\
     \footnotesize perspectivist annotations\\
     \footnotesize first-order labels \circled{1}\circled{2}};
 
\node[gptbox, right=16mm of mapinput, minimum width=40mm] (extract)
    {\textbf{GPT-5 Extraction} (\S\ref{sec:tom-module})\\[2pt]
     \circled{1}--\circled{2}: \citep{li2026groundedmisunderstandingsasymmetricdialogue} first-order labels\\
     \circled{3}--\circled{4}: inferred second-order beliefs
     
     + 3 auxiliary fields};
 
\coordinate (hvpos) at ($(liinput.south -| extract) + (0,5mm)$);
\node[humanbox, at=(hvpos), anchor=north,
      minimum width=40mm, minimum height=13mm] (human)
    {\textbf{Human Validation}\\[2pt]
     Blind annotation pilot ($n{=}60$)\\
     $\kappa = 0.80$ (speaker ToM)\\
     $\kappa = 0.77$ (addressee ToM)};
 
\node[rulebox, right=8mm of extract, minimum width=28mm, minimum height=22mm] (phi)
    {\textbf{Friction Module $\Phi$} (\S\ref{sec:friction-derivation})\\[2pt]
     \footnotesize (deterministic, rule-based)\\[3pt]
     $F^{-} = \sum_i w_i\, C_i$\\[2pt]
     $F^{+} = \text{INFOGAIN}(h,a)$};
 
\node[dpobox, right=34mm of phi, yshift=32mm] (dpo)
    {\textbf{DPO}\\ \footnotesize baseline};
\node[farbox, below=10mm of dpo] (far)
    {\textbf{FAR}\\ \footnotesize reward shaping\\ \footnotesize $R' = R_{\text{task}} + \alpha\, g(\text{Risk})F - \beta C_{\text{fric}}$};
\node[fppbox, below=6mm of far] (fpp)
    {\textbf{FPP}\\ \footnotesize preference pairing\\ \footnotesize rank by $F^{+}(y^+)\!>\!F^{+}(y^-)$};
\node[ftrbox, below=6mm of fpp] (ftr)
    {\textbf{FTR}\\ \footnotesize trust-region control\\ \footnotesize $\beta(h)=\dfrac{1}{\epsilon_0+\kappa F^{-}(h)}$};
 
\begin{scope}[on background layer]
\node[groupbox={FPO Trainer Variants (\S\ref{sec:trainer-adaptation})}, fit=(far)(fpp)(ftr)] (grp) {};
\end{scope}
 
\node[inputbox, right=8mm of grp, yshift=0mm, minimum height=14mm] (out)
    {Trained policy (\S\ref{sec:downstream})\\ \footnotesize (evaluated on held-out\\ \footnotesize MapTask test set)};
 
\draw[arr] (mapinput) -- (extract);
\draw[arr] (liinput.east) -- ($(liinput.east)+(6mm,0)$) |- (extract.west)
    node[pos=0.15, above, font=\scriptsize] {provides \circled{1}\circled{2}};
\coordinate (hv1) at ($(mapinput.south)+(0,-3mm)$);
\coordinate (hvL) at ($(liinput.south west)+(-4mm,0)$);
\coordinate (hv2) at (hvL |- hv1);
\coordinate (hv3) at (hvL |- human.west);
\draw[arr] (mapinput.south) -- (hv1) -- (hv2) -- (hv3) -- (human.west)
    node[pos=0.85, above, font=\scriptsize] {sample $n{=}60$};
\draw[arr, dashed] (extract.south) -- (human.north)
    node[midway, right, font=\scriptsize] {audits};
\draw[arr] (extract) -- (phi);
 
\coordinate (branch) at ($(phi.east)+(6mm,0)$);
\draw[line width=0.7pt] (phi.east) -- (branch);
 
\draw[arr] (mapinput.north) |- (dpo.west)
    node[pos=0.7, above, font=\scriptsize] {no per-instance $F^{-}$ input};
\draw[line width=0.7pt] (branch) -- (branch |- far.west);
\draw[arr] (branch |- far.west) -- (far.west)
    node[midway, above, font=\scriptsize] {$F^{-}$ inject};
 
\draw[line width=0.7pt] (branch) -- (branch |- fpp.west);
\draw[arr] (branch |- fpp.west) -- (fpp.west)
    node[midway, above, font=\scriptsize] {$F^{+}$ ranking};
 
\draw[line width=0.7pt] (branch) -- (branch |- ftr.west);
\draw[arr] (branch |- ftr.west) -- (ftr.west)
    node[midway, above, font=\scriptsize] {$F^{-}$ trust region};
 
\coordinate (busx) at ($(grp.east)!0.55!(out.west)$);
\draw[line width=0.7pt] (dpo.east) -- (busx |- dpo.east);
\draw[line width=0.7pt] (far.east) -- (busx |- far.east);
\draw[line width=0.7pt] (fpp.east) -- (busx |- fpp.east);
\draw[line width=0.7pt] (ftr.east) -- (busx |- ftr.east);
\draw[line width=0.7pt] (busx |- dpo.east) -- (busx |- ftr.east);
\draw[arr] (busx |- out.west) -- (out.west);

\end{tikzpicture}%
}
\caption{Pipeline overview. GPT-5 augments \citet{li2026groundedmisunderstandingsasymmetricdialogue}’s first-order labels with second-order ToM predictions and auxiliary fields to construct the four-identifier schema. A blind human-annotation pilot audits the inferred second-order labels. The deterministic friction module $\Phi$ computes $F^-$ and $F^+$,
which condition the FAR, FPP, and FTR objectives. Standard DPO uses friction-informed preference pairs but receives no per-instance friction value; DPO-FB removes friction from pair construction entirely.}
\label{fig:pipeline}
\end{figure*}

\section{Introduction}
\label{sec:introduction}

Recent work on epistemically asymmetric interaction has begun to expose a structural gap in how collaborative dialogue is modeled. In settings such as the Distributed Partial Information Puzzle \citep[DPIP;][]{zhu2026distributedpartialinformationpuzzles}, each participant observes only a partial view of the environment, requiring inference not only about what others know, but also about what they can perceive and misinfer. Such interactions reveal a problem that standard turn-level evaluation obscures: interlocutors may appear aligned through acknowledgments, uninterrupted task progression, or lexical accommodation while still grounding expressions in different referents or maintaining incompatible interpretations. We refer to this gap as the distinction between \textit{surface-level coordination} (observable dialogue behavior) and \textit{epistemic alignment} (convergence of participants' belief states, including beliefs about others' beliefs); alignment is therefore better viewed as ongoing regulation of epistemic states rather than as successful surface coordination.


Frictive Policy Optimization \citep[FPO;][]{pustejovsky2026frictivepolicyoptimizationllms} addresses this distinction by formalizing alignment as risk-sensitive control over belief, commitment, and uncertainty rather than turn-level optimization against a static preference signal. Existing instantiations, however, primarily operate on surface-level or indirectly represented frictive states, limiting their ability to capture latent divergence in genuinely asymmetric interaction. Such divergence can require an explicit account of the higher-order beliefs:  participants may hold incompatible first-order interpretations while each presumes, at the second-order level, that the other shares their own. We therefore extend FPO with a point-in-time Theory-of-Mind (ToM) representation that tracks, for each referring expression, the speaker's intended referent, the addressee's interpretation, and each participant's model of the other's belief. Friction is then computed from this epistemic structure rather than inferred from dialogue acts alone. 


We make three contributions. First, we introduce a point-in-time ToM schema for HCRC MapTask \citep{anderson1991hcrc} that reconstructs first- and second-order beliefs at each referring expression and converts their divergence into a computable friction signal. Second, channel attribution shows that second-order belief structure materially improves recovery of misunderstandings, including \emph{silent intent fixing}. Third, across FAR and FTR, per-instance friction conditioning improves intervention behavior and warranted-context calibration over DPO while substantially increasing training stability. 

\vspace{-.5em}

\section{Background}
\label{sec:background}
\vspace{-.5em}
\subsection{LLM Alignment}
\label{sec:bg-alignment}
\vspace{-.5em}
Mainstream alignment methods---RLHF \citep{NIPS2017_d5e2c0ad}, DPO \citep{rafailov2023direct}, and GRPO \citep{shao2024deepseekmathpushinglimitsmathematical}---optimize policies toward turn-level preference signals. This paradigm is well suited to instruction following but does not explicitly represent the evolving epistemic states that determine response appropriateness in collaborative interaction. Frictive Policy Optimization \citep[FPO;][]{pustejovsky2025frictive,pustejovsky2026frictivepolicyoptimizationllms} instead formulates alignment as risk-sensitive epistemic control through a friction functional.
Its prior implementation has derived friction from natural-language descriptions of frictive states \citep{nath-etal-2025-frictional} and simulated partial-information environments \citep{nath2026craftgroundedmultiagentcoordination}; whereas, how to compute it directly from participants' belief structure in naturalistic asymmetric dialogue remains open.
\vspace{-.5em}
\subsection{Grounding, Perspective, and Epistemic Alignment}
\label{sec:bg-tom}

Common ground is dynamically established through grounding \citep{Clark1991-CLABG-2,clark1996using}, yet participants may proceed as if mutual understanding has been achieved while maintaining divergent interpretations \citep{zhu2026propositional}. Work on automated common ground tracking~\cite{khebour-etal-2024-common} shows that surface transcripts alone are insufficient even for tracking shared beliefs in collaborative tasks, underscoring the need for richer epistemic representations.



Theory of Mind (ToM) distinguishes first-order beliefs from beliefs about others' beliefs \citep{premack1978does,wimmer1983beliefs,sperber1986relevance}, making such silent divergence representable. \citet{quesque2020theory} argue that genuine mentalizing requires both representing another's mental state rather than relying only on behavioral or textual regularities and maintaining that perspective as distinct from one's own. \citet{pmlr-v267-riemer25a} distinguish \emph{literal ToM}, predicting another agent's behavior, from \emph{functional ToM}, adapting one's own behavior using a representation of another's mental state. Our use of ToM follows this functional view: epistemic representations are constructed not only to predict beliefs, but to condition downstream intervention.


Recent benchmarks probe these capabilities in increasingly interactive settings. FANToM \citep{kim2023fantom} evaluates belief attribution and information asymmetry in multi-party dialogue; MMToM-QA \citep{jin2024mmtom} requires joint reasoning over visual and linguistic evidence; and NegotiationToM \citep{chan2024negotiationtom} evaluates belief and intention inference during strategic negotiation. These benchmarks primarily evaluate reasoning performance over annotated mental states. Our focus is complementary: representing epistemic relations explicitly so that their divergence can be converted into a downstream control signal.

\vspace{-.5em}
\section{Data}
\label{sec:data}
\vspace{-.5em}

We construct FPO training instances from \citet{li2026groundedmisunderstandingsasymmetricdialogue}, which augments the HCRC MapTask corpus with perspectivist grounding annotations. In MapTask, a \emph{giver} guides a \emph{follower} through a navigation task using mismatched maps. Giver/follower roles remain fixed across a dialogue, whereas \emph{speaker} and \emph{addressee} vary with each referring expression (RE), which determines the assignment of first- and second-order beliefs (\S\ref{sec:tom-module}). The annotations distinguish three map discrepancies: \emph{existence} (a landmark appears on only one map), \emph{lexical} (the same landmark has different labels), and \emph{multiplicity} (a landmark is duplicated on one map only), with multiplicity providing the main source of silent misalignment.



Each RE is aligned with its perspectivist belief annotations and yields an instance containing the dialogue history, giver/follower beliefs, and a grounding status (\textsc{pending}, \textsc{misunderstood}, or \textsc{aligned}). We treat \textsc{pending} and \textsc{misunderstood} instances as frictive for friction construction, while \textsc{aligned} instances provide non-frictive contrasts. The resulting split contains 99 training dialogues with 2,705 REs (147 misunderstood) and 24 test dialogues with 761 REs (69 misunderstood), preserving misunderstanding rates of 5.5\% and 9.1\%, respectively. After the additional filtering used for policy adaptation, 663 training and 278 held-out evaluation instances remain. Details about data preprocessing and split are in Appendix~\ref{app:data-preprocessing}.

\begin{table*}[t]
\centering
\small
\begin{tabularx}{\textwidth}{lcccX}
\toprule
Configuration & \textcircled{\small 1}-\textcircled{\small 2} & \textcircled{\small 3}-\textcircled{\small 1} & \textcircled{\small 4}-\textcircled{\small 2} & Description \\
\midrule
Aligned                      & $=$    & $=$    & $=$    & Both parties land on the same landmark and correctly believe they have. \\[0.4em]
Silent intent fixing         & $\neq$ & $=$    & $=$    & Divergence exists but neither party detects it; both proceed confidently. \\[0.4em]
Asymmetric (speaker-aware)   & $\neq$ & $\neq$ & $=$    & Divergence exists; speaker has detected it, addressee has not. \\[0.4em]
Asymmetric (addressee-aware) & $\neq$ & $=$    & $\neq$ & Divergence exists; addressee has detected it, speaker has not. \\[0.4em]
Mutual awareness             & $\neq$ & $\neq$ & $\neq$ & Divergence exists and both parties have detected it; repair is imminent or underway. \\
\bottomrule

\end{tabularx}
\caption{The five canonical epistemic configurations produced by the
four-identifier schema. Columns report pairwise agreement among the four identifiers (\S\ref{sec:tom-module}):
\textcircled{\small 1}-\textcircled{\small 2} = referent agreement between speaker and addressee (first-order);
\textcircled{\small 3}-\textcircled{\small 1} = whether the speaker correctly anticipates the addressee's interpretation;
\textcircled{\small 4}-\textcircled{\small 2} = whether the addressee correctly recovers the speaker's intent.
``$=$'' is agreement, ``$\neq$'' is divergence.}
\label{tab:four-configs}
\vspace{-1em}
\end{table*}

\vspace{-.5em}

\section{Theory-of-Mind-Grounded Friction Construction}
\label{sec:pipeline}
\vspace{-.5em}

    We adopt the functional conception of ToM proposed by \citet{pmlr-v267-riemer25a}: the value of a mental-state representation lies not in predicting another agent's behavior, but in guiding one's own. Accordingly, our friction signal conditions policy intervention on an explicitly represented divergence between an agent's own belief and its model of the interlocutor's. This design satisfies the two requirements identified by \citet{quesque2020theory}: it operates over explicit mental-state representations rather than surface behavioral regularities, and it maintains distinct perspectives by computing friction from the comparison between an agent's belief and its separately represented model of the addressee's belief.
 
Figure~\ref{fig:pipeline} gives an overview of the full pipeline described
in this section. GPT-5~\cite{singh2026openaigpt5card} extracts the four-identifier belief structure
(\S\ref{sec:tom-module}) from dialogue and map context, validated
against a blind human-annotation pilot; the deterministic friction module
$\Phi$ (\S\ref{sec:friction-derivation}) then converts this structure into
$F^-$ and $F^+$ signals, which are injected into three FPO trainer variants
(FAR, FPP, and FTR alongside a
friction-blind DPO baseline for comparison (\S\ref{sec:trainer-adaptation}).

\vspace{-.5em}

\subsection{Point-in-Time ToM Extraction}
\label{sec:tom-module}

We extend the perspectivist annotation framework of 
\citet{li2026groundedmisunderstandingsasymmetricdialogue} in two ways: (1) replacing retrospective 
grounding outcomes with point-in-time belief attribution, and (2) 
augmenting first-order beliefs with explicit second-order 
Theory-of-Mind (ToM) beliefs. The resulting representation
makes silent divergence detectable through cross-party 
belief comparison.


We therefore introduce a four-identifier schema that attributes, for 
each referring expression at time $t$: \textcircled{\small1} the 
speaker's intended referent, \textcircled{\small2} the addressee's 
inferred referent, \textcircled{\small3} the speaker's belief about 
the addressee's interpretation, and \textcircled{\small4} the 
addressee's belief about the speaker's intention.  The first-order identifiers \textcircled{\small1}--\textcircled{\small2} come from the perspectivist annotations of \citet{li2026groundedmisunderstandingsasymmetricdialogue}; the second-order identifiers \textcircled{\small3}--\textcircled{\small4} are inferred by GPT-5 from dialogue, map, and dialogue-act context. The attributed state is indexed to time $t$, although the offline extractor may use later repair or contradiction evidence to reconstruct that latent state. Silent divergence therefore becomes a relation among participant-specific beliefs rather than a single grounding label.



Beyond the four identifiers, the schema extracts three auxiliary fields through GPT-5 inference: \textsc{uptake\_quality}, \textsc{surface\_signals},  and \textsc{is\_existence\_query}.

\textsc{uptake\_quality} is a categorical field in the \texttt{addressee\_belief} block with four values: \texttt{committed}, \texttt{hesitant}, \texttt{withheld}, and \texttt{absent}. It captures how confidently the addressee responds after an RE, ranging from confident acknowledgment to explicit non-commitment or no response. The friction module maps these categories to the uncertainty component of $F^{-}$, treating withholding as the strongest signal of unresolved uncertainty.

\textsc{surface\_signals} is a multi-label field recording repair-oriented discourse markers in the addressee response, including \texttt{acknowledge}, \texttt{pause}, \texttt{request\_clarify}, \texttt{repair}, and \texttt{contradict}. The friction module uses these signals both to estimate the contradiction component of $F^{-}$ and to infer whether the addressee is aware of a grounding mismatch.

\textsc{is\_existence\_query} is a boolean field in the \texttt{speaker\_belief} block identifying references that explicitly ask whether a landmark exists (e.g., ``do you have a forest?''). These cases are excluded from FPO training because discovering asymmetry through questioning is treated as successful information exchange rather than grounding failure.

Pairwise comparisons among the four identifiers induce five canonical 
epistemic configurations shown in Table~\ref{tab:four-configs}\footnote{Two pairwise comparisons are not shown, \circled{1}-\circled{4} and
\circled{2}-\circled{3}. These are not independent degrees of freedom: \circled{1}-\circled{4}
is entailed by the reported columns whenever \circled{4}-\circled{2}=``='', and analogously
\circled{2}-\circled{3} is entailed whenever \circled{3}-\circled{1}=``=''; both are therefore
fixed in the Aligned and Silent-intent-fixing rows and partially fixed in the two
singly-aware rows, leaving them genuinely underdetermined only in the Mutual-awareness
configuration. Thus, they do not enter the friction computation.}. 
\emph{Aligned} is the only non-frictive configuration, while 
\emph{silent intent fixing}, as the central friction target, captures 
cases where interlocutors anchor to different referents while mutually 
presuming agreement.

\paragraph{Illustrative example}
Consider a multiplicity discrepancy in which the giver's map contains two
instances of the same landmark while the follower's map contains only one.
When the giver refers to one of the two instances, the follower may resolve
the expression to their single corresponding landmark. The intended and
interpreted referents therefore diverge
(\circled{1}$\neq$\circled{2}), while the giver assumes the follower recovered
the intended referent (\circled{3}=\circled{1}) and the follower assumes their
interpretation matches the giver's intention
(\circled{4}=\circled{2}). This yields \emph{silent intent fixing}:
first-order interpretations diverge while neither participant represents the
mismatch.

\vspace{-.5em}
\subsection{Friction Computation}
\label{sec:friction-derivation}



The friction module $\Phi$ maps the four-identifier schema to FPO's productive friction signal ($F^-$) and unproductive friction signals ($F^+$): {\small \begin{alignat}{2} &F^-(h) = w_{\textsc{Unc}}\,\textsc{Unc}(h) + w_{\textsc{Contr}}\,\textsc{Contr}(h) \nonumber\\ &\hspace{4em} + w_{\textsc{Haz}}\,\textsc{Haz}(h) + w_{\textsc{ValConf}}\,\textsc{ValConf}(h) \label{eq:f-minus-composite}\\[0.3em] &F^+(h,a)=\textsc{InfoGain}(h,a). \label{eq:f-plus-table} \end{alignat} }

All $F^-$ components are deterministic scalar surrogates computed from the four-identifier schema and auxiliary dialogue-observable fields,
normalized to $[0,1]$. \textsc{Unc} captures uncertainty from addressee uptake; \textsc{Contr} captures surfaced contradiction or repair; \textsc{Haz} estimates the propagation risk associated with map asymmetries; and \textsc{ValConf} captures second-order belief conflict from the four-identifier structure. In particular, referential divergence is defined by $\textcircled{\small1}\neq\textcircled{\small2}$, while speaker and addressee awareness are reflected by $\textcircled{\small3}\neq\textcircled{\small1}$ and $\textcircled{\small4}\neq\textcircled{\small2}$, respectively. Thus, $F^-$ is not simply an agreement/disagreement polarity score: it combines uncertainty, surfaced repair, propagation risk, and higher-order belief conflict. Full component mappings appear in Appendix~\ref{app:f-}.




\paragraph{Composition.} We use $w$ = (0.15, 0.20, 0.30, 0.35) for $(\textsc{Unc},\textsc{Contr},\textsc{Haz},\textsc{ValConf})$, respectively, with $\sum_i w_i=1$. The weights are hand-set rather than fit to this dataset. Greater mass is assigned to \textsc{Haz} and \textsc{ValConf}, which target propagation-prone and potentially latent misalignment, whereas \textsc{Unc} and \textsc{Contr} primarily reflect already-surfaced precursor signals. We do not claim these values are optimal: uniform weighting preserves the relevant friction ordering, and $\pm20\%$ perturbations do not change the qualitative downstream conclusions (Appendix~\ref{app:sensitivity}).

\paragraph{Productive friction.} $F^+(h,a)$ represents the epistemic utility of an intervention over the four actions \texttt{clarify}, \texttt{verify}, \texttt{redirect}, and \texttt{refuse}, inherited from the FPO action space in \citet{pustejovsky2026frictivepolicyoptimizationllms} rather than proposed here as a discourse taxonomy. In our offline implementation, $\textsc{InfoGain}(h,a)$ is a hand-specified lookup over epistemic configuration and action: diagnostic actions receive greater utility in divergent states, whereas unnecessary intervention receives little value in aligned states. The full utility table are reported in Appendix~\ref{app:f+}.

\vspace{-.5em}
\section{Friction Signal Validation}
\label{sec:validation}
\vspace{-.5em}
Before using the friction cache for trainer adaptation (\S\ref{sec:trainer-adaptation}), we validate the signal along three axes: (i) whether the schema recovers usable belief structure and misunderstanding labels, (ii) whether the composite $F^-$ monotonically separates frictive from non-frictive contexts, and (iii) whether the signal materially depends on the second-order ToM structure introduced in \S\ref{sec:tom-module}. Results are reported on both train and test partitions below.
\vspace{-.5em}
\subsection{Schema Validation and Belief Accuracy}
\label{sec:validation:schema}

The four-identifier schema achieves $98.5\%$ parse success on train ($2665/2705$) and $100\%$ on test ($761/761$); the remaining train failures are primarily API truncations due to long-dialogue and are filtered before downstream training. Belief accuracy against \citet{li2026groundedmisunderstandingsasymmetricdialogue} is shown in Table~\ref{tab:belief-acc}. Addressee-side belief recovery substantially exceeds speaker-side recovery ($80.6$--$81.3\%$ vs.\ $59.8$--$64.1\%$ loose match), which is expected because addressee interpretations are more directly observable through uptake and repair behavior, whereas speaker intent must be inferred from production-side evidence alone. Since both friction computation and trainer adaptation primarily condition on addressee-side state, the stronger follower-side accuracy is the operationally relevant quantity.

Second-order ToM consistency is lower ($35.9$--$43.5\%$ speaker ToM; $41.4$--$42.0\%$ addressee ToM), but these values are reported as soft signals rather than gold-aligned accuracy because no direct annotation for second-order beliefs exists in the corpus. To directly assess the reliability of GPT-5 second-order predictions, we conducted a stratified blind manual annotation on 60 samples (details in Appendix \ref{app:human-annotation}). This independent human annotation then was used as ground-truth that the GPT-5 predictions were compared against. Agreement reached $82\%$ for speaker-ToM and $80\%$ for addressee-ToM, corresponding to Cohen's $\kappa$ values of $0.80$ and $0.77$, respectively. These results indicate substantial agreement with human judgments, supporting the use of the inferred second-order beliefs as supervisory signals during training.

\begin{table}[t]
\centering
\small
\begin{tabular}{@{}lrr@{}}
\toprule
Quantity (loose match) & Train & Test \\
\midrule
Speaker belief        & 59.8\% & 64.1\% \\
Addressee belief      & 81.3\% & 80.6\% \\
Speaker ToM consistency   & 35.9\% & 43.5\% \\
Addressee ToM consistency & 41.4\% & 42.0\% \\
\bottomrule
\end{tabular}
\caption{Four-identifier accuracy on train and test partitions.}
\label{tab:belief-acc}
\vspace{-1em}
\end{table}

At the task level, the schema recovers misunderstanding labels with $67.6\%$ recall on train ($98/145$) and $65.2\%$ on test ($45/69$), improving more than $60$ points over the cascade-based alternatives discussed in Appendix~\ref{app:tom-extraction}.
\vspace{-.5em}
\subsection{$F^-$ Distribution and Monotonic Ordering}
\label{sec:validation:separation}

The composite $F^-$ is intended to function as a monotonic risk signal: genuinely frictive contexts should receive systematically higher values than non-frictive ones. This ordering holds consistently across both partitions (Table~\ref{tab:friction-distribution}). Mean $F^-$ increases from $0.185$ to $0.375$ on train ($\Delta=0.190$) and from $0.193$ to $0.390$ on test ($\Delta=0.197$) when moving from pending to misunderstood contexts. The cross-partition gap appears minimal, as we argue the composite generalizes across the data. This monotonicity is required for downstream trainer adaptation, where $F^-(h)$ directly controls trust-region width: a non-monotonic signal would incorrectly relax constraints on high-risk contexts or tighten them on benign ones.

\begin{table}[t]
\centering
\small
\begin{tabular}{@{}llrrr@{}}
\toprule
Partition & Status & $n$ & Mean $F^-$ & Median \\
\midrule
\multirow{2}{*}{Train}
 & Pending       & 2521 & 0.185 & 0.182 \\
 & Misund        &  145 & 0.375 & 0.340 \\
\midrule
\multirow{2}{*}{Test}
 & Pending       &  692 & 0.193 & 0.203 \\
 & Misund        &   69 & 0.390 & 0.340 \\

\bottomrule
\end{tabular}
\caption{$F^-$ distribution by gold status on both partitions}
\label{tab:friction-distribution}
\vspace{-1.5em}
\end{table}
\subsection{Detection Channels and ToM Contribution}
\label{sec:validation:channels}

We further decompose detected misunderstandings by the channel responsible for recovery: (i) ToM-dependent configurations involving silent intent fixing or asymmetric awareness, (ii) direct first-order divergence, and (iii) metadata-based multiplicity fallback. Results are summarized in Table~\ref{tab:misund-channels}.


\begin{table}[t]
\centering
\small
\begin{tabular}{@{}lrr@{}}
\toprule
Detection channel & Train (\%) & Test (\%) \\
\midrule
ToM-required configurations & 44.2 & 39.1 \\
First-order divergence & 1.4 & -- \\
Multiplicity fallback & 22.1 & 26.1 \\
\midrule
Total recall & 67.6 & 65.2 \\
\bottomrule
\end{tabular}
\caption{Channel decomposition of detected misunderstandings.}
\label{tab:misund-channels}
\vspace{-1.5em}
\end{table}


Second-order structure accounts for a substantial share of recovered misunderstandings, including all silent-intent-fixing instances identified by our framework. Removing the ToM-dependent channel reduces recall from $67.6\%$ to $23.4\%$ on train and from $65.2\%$ to $26.1\%$ on test, while removing the multiplicity fallback as well reduces recall to $1.4\%$ and $0\%$, respectively. The channels are therefore non-redundant, and the sharp drop under ToM ablation shows that the friction signal depends materially on the second-order belief structure introduced in \S\ref{sec:tom-module}. Most remaining misses involve underspecified or visually ambiguous references that the upstream extractor cannot disambiguate from dialogue and map evidence alone.

\vspace{-.5em}
\section{FPO Trainer Adaptation}
\label{sec:trainer-adaptation}
\vspace{-.5em}

We adapt a friction-blind DPO baseline \citep{rafailov2023direct} and
three FPO variants (FAR, FPP, and FTR) from
\citet{pustejovsky2026frictivepolicyoptimizationllms}  to offline
MapTask data. We use the participant-agent setting: the policy observes
only its own map-derived information and dialogue history and decides
whether to \texttt{answer} or intervene. Full cross-party ToM annotations
are used offline to construct friction supervision but are never direct
policy inputs. During both training and standard inference, the policy
receives dialogue history plus the same compact agent-observable epistemic
projection. 


\paragraph{Training instances and observable belief input.}
Each instance pairs a referring expression with its immediate addressee
response. The policy input consists of the dialogue history and an
\emph{observable belief-state preamble}: a compact projection of the
schema-derived epistemic state containing only information available to the
participant-agent. Full cross-party ToM annotations and the other
participant's private map are used only offline to construct friction
supervision and are never direct policy inputs. The same observable input is
used during training and standard inference. After filtering underspecified
and existence-query cases, the policy dataset contains 663 training instances
(216 divergent, 447 aligned) and 278 held-out test instances. For pair-based methods, completions are retrieved from the corpus: repair is
preferred over acknowledgment in divergent contexts, whereas unnecessary
intervention is dispreferred in aligned contexts.


\paragraph{DPO.}
Standard DPO uses $\beta=0.1$ with a frozen
reference policy and receives no friction signal.

\paragraph{FAR (Friction-Aware Reward).}

FAR incorporates friction through reward shaping,
\[
R' = R_{\text{task}}
+\alpha\,g(\mathrm{Risk})F(h,y)
-\beta\,C_{\text{fric}}(a),
\]
so intervention is encouraged when epistemic risk is high and penalized when
it imposes unnecessary cost. We instantiate $\mathrm{Risk}$ from the
friction-derived epistemic state; offline adaptation details are given in
Appendix~\ref{app:far}.

\paragraph{FPP (Friction Preference Pairing).}
FPP applies a DPO-style preference objective in which $y^+$ must have greater
productive friction $F^+$ than $y^-$. This converts the friction signal into
pairwise preference supervision while retaining the same underlying policy
backbone. The offline pair-construction and weighting adjustments are
described in Appendix~\ref{app:fpp}.
 
\paragraph{FTR (Friction-Conditioned Trust Region).}

FTR uses epistemic risk to control how far the policy may depart from the base
model: high-friction contexts permit larger updates, while low-friction
contexts remain more strongly anchored. We implement
\begin{align*}
L_{\text{FTR}}(h) &= L_{\text{CE}}(y_{\text{gold}}|h)
    + \beta(h)\widehat{\mathrm{KL}}(\pi_\theta\|\pi_0|h) \\\beta(h) &= \frac{1}{\epsilon_0+\kappa F^-(h)}.
\end{align*}
Estimator and tuning details appear in Appendix~\ref{app:ftr}.

\section{Evaluation}
\label{sec:downstream}

We evaluate three questions: whether per-instance ToM-grounded friction
improves policy adaptation over static preference training; whether the gains
come from friction-aware supervision or direct objective conditioning; and
whether they persist under prompting baselines and removal of inference-time
belief information. All experiments use the held-out MapTask test set
($n{=}278$), with 120 contexts where intervention is warranted and 158
aligned contexts.

Our primary measures are intervention F1 and calibration on warranted
contexts. ClarifyScore/F1 rewards intervention under epistemic
divergence while penalizing unnecessary intervention on aligned instances.
Calibration is evaluated with both ECE (expected binned gap between
first-token confidence and intervention correctness), and the
Brier score, a strictly proper scoring rule. We report Brier score
as an independent check that calibration conclusions do not depend on ECE
alone. InfoEff, expected epistemic gain per unit intervention cost,
and pairwise LLM preference provide secondary evidence.

Primary intervention labels are assigned by a rule-based classifier rather
than an LLM. Warrant labels are derived from the held-out four-identifier
annotations introduced in \S\ref{sec:tom-module}, so the principal results
do not depend on LLM judging of generated responses. Pairwise comparisons
use Gemma-4-31B-IT \citep{gemmateam2026gemma4}, evaluating each pair in both orders
to reduce position bias. 

\subsection{Main Results and Training Stability}
\label{sec:downstream:headline}

We compare DPO with three FPO adaptations: reward shaping (FAR),
friction-based preference pairing (FPP), and friction-conditioned
trust-region control (FTR). Table~\ref{tab:downstream-headline} reports
the primary-run results, and Table~\ref{tab:multiseed} summarizes
stability across three independent training runs.

\begin{table*}[t]
\centering
\small
\setlength{\tabcolsep}{4pt}
\begin{tabular}{lcccccc}
\toprule
Policy & F1 & ECE$_{\mathrm{warr}}$ & ECE$_{\mathrm{align}}$ &
Brier$_{\mathrm{warr}}$ & InfoEff & Net win vs.\ DPO \\
\midrule
DPO
& $0.344$
& $0.432$
& $\mathbf{0.148}$
& $0.411$
& $\mathbf{3.99}$
& --- \\
FAR
& $0.416$
& $0.227$
& $0.208$
& $\mathbf{0.281}$
& $3.92$
& $+15.8$ \\
FPP$^\dagger$
& $0.603$
& $0.000$
& $1.000$
& $0.000$
& $3.78$
& $-91.4$ \\
FTR
& $\mathbf{0.417}$
& $\mathbf{0.178}$
& $0.220$
& $0.293$
& $3.94$
& $+13.3$ \\
\bottomrule
\end{tabular}
\caption{Primary-run evaluation on the held-out test set
($n{=}278$; 120 warranted and 158 aligned contexts). 95\% bootstrap CIs (B=2,000).
Net win is paired win-minus-loss percentage against DPO under
Gemma-4-31B-IT.}
\label{tab:downstream-headline}
\end{table*}

FAR and FTR improve intervention F1 over DPO while substantially reducing miscalibration where intervention is warranted: ECE falls from $0.432$ to $0.227$ and $0.178$, respectively. Brier score independently confirms this result, decreasing from $0.411$ to $0.281$ and $0.293$. As a strictly proper scoring rule, Brier score shows that the calibration gain is not an artifact of ECE binning or an FPO-specific metric. FAR and FTR incur only a modest calibration tradeoff on aligned contexts, while InfoEff remains comparable, indicating that their gains do not arise from intervening indiscriminately.

\paragraph{Stability across three training runs.}
We repeat the full training and evaluation procedure from three independent QLoRA~\cite{dettmers2023qloraefficientfinetuningquantized} initializations (seeds $42$, $123$, and $7$). 
Across three independent QLoRA initializations, DPO is substantially more
variable than FAR and FTR: its F1 ranges from $0.152$ to $0.398$
(std.\ $0.106$), compared with standard deviations of $0.023$ and $0.009$
for FAR and FTR, respectively. All six same-seed FAR/FTR comparisons against
DPO are significant under McNemar tests ($p \le 3\times10^{-3}$; full
per-seed results in Appendix~\ref{app:multiseed}).

\begin{table}[t]
\centering
\small
\setlength{\tabcolsep}{6pt}
\begin{tabular}{lcc}
\toprule
Method & F1 & ECE$_{\mathrm{warr}}$ \\
\midrule
DPO
& $0.298 \pm 0.106$
& $0.523 \pm 0.090$ \\
FAR
& $0.411 \pm 0.023$
& $0.225 \pm 0.038$ \\
FTR
& $\mathbf{0.430 \pm 0.009}$
& $\mathbf{0.192 \pm 0.011}$ \\
\bottomrule
\end{tabular}
\caption{Mean $\pm$ standard deviation across three independent
training runs. Per-seed results appear in
Appendix~\ref{app:multiseed}.}
\vspace{-1em}
\label{tab:multiseed}

\end{table}

\paragraph{Failure of preference-only friction.}
FPP obtains superficially high F1 by intervening on every test instance. Its perfect warranted-context scores follow mechanically from perfect recall, while its aligned-context ECE of $1.0$ and strongly negative judge preference expose the collapse. Because its pairwise objective provides no absolute signal for when intervention is unnecessary, this result shows that friction must regulate behavior at the instance level rather than only rank completions.

\paragraph{Secondary judge evidence.}
LLM pairwise judgments favor FAR and FTR in the full sample, but the
differences become inconclusive after response-length control. We therefore
treat judge preference as secondary to rule-based F1 and calibration.

\subsection{Sources of the Improvement}
\label{sec:downstream:sources}

We use staged controls to separate three potential sources of the downstream
gains: friction-aware supervision, direct conditioning on per-instance
friction, and the second-order ToM structure used to compute that friction.

\paragraph{Friction-aware supervision and objective conditioning.}
We compare three levels of friction use. DPO-FB is fully friction-blind:
the chosen completion is the observed corpus response and the rejected
completion is sampled from another dialogue without reference to epistemic
configuration. Standard DPO uses friction-informed response pairs but does
not receive the per-instance friction value. FAR and FTR additionally
condition their objectives directly on that value.

As shown in Table~\ref{tab:friction-dose}, the staged progression increases intervention F1 from $0.282$ for DPO-FB
to $0.344$ for DPO and $0.416$--$0.417$ for FAR/FTR. Warranted-context
Brier error likewise decreases from $0.532$ to $0.411$ and then to
$0.281$--$0.293$. This pattern is consistent with two separable
contributions: friction-informed supervision improves over friction-blind
pairing, while direct per-instance conditioning provides an additional gain.
\begin{table}[t]
\centering
\small
\setlength{\tabcolsep}{5pt}
\begin{tabular}{lccc}
\toprule
Condition & F1 & ECE & Brier$_{\mathrm{warr}}$ \\
\midrule
DPO-FB & $0.282$ & $0.283$ & $0.532$ \\
DPO    & $0.344$ & $0.174$ & $0.411$ \\
FAR    & $0.416$ & $\mathbf{0.152}$ & $\mathbf{0.281}$ \\
FTR    & $\mathbf{0.417}$ & $0.189$ & $0.293$ \\
\bottomrule
\end{tabular}
\caption{Staged friction controls under the primary-run evaluation
protocol. DPO-FB removes friction from pair construction; DPO uses
friction-informed pairs without per-instance objective conditioning;
FAR and FTR incorporate the friction value directly. ECE is global;
Brier score is reported on warranted contexts.}
\label{tab:friction-dose}
\vspace{-1.5em}
\end{table}

\paragraph{Policy-level contribution of second-order ToM.}
To isolate the proposed ToM extension, we train lower-order FAR and FTR
variants with the same data and objectives but compute friction using only
surface and first-order channels. This removes second-order configurations
such as silent intent fixing and asymmetric awareness while preserving the
remaining training setup.

\begin{table}[t]
\centering
\footnotesize
\setlength{\tabcolsep}{3pt}
\resizebox{\columnwidth}{!}{%
\begin{tabular}{llccccc}
\toprule
Objective & Friction & F1 & $\Delta$F1 & Rel.\ gain & ECE & Brier$_{\mathrm{warr}}$ \\
\midrule
FAR & Lower-order
& $0.368$
& ---
& ---
& $0.270$
& $\mathbf{0.253}$ \\
& Full ToM
& $\mathbf{0.416}$
& $+0.048$
& $\mathbf{+13.0\%}$
& $\mathbf{0.152}$
& $0.281$ \\
\midrule
FTR & Lower-order
& $0.400$
& ---
& ---
& $\mathbf{0.135}$
& $\mathbf{0.278}$ \\
& Full ToM
& $\mathbf{0.417}$
& $+0.017$
& $\mathbf{+4.3\%}$
& $0.189$
& $0.293$ \\
\bottomrule
\end{tabular}
}
\caption{Primary-run policy-level ablation of second-order ToM.
Lower-order variants retain surface and first-order friction channels
but remove second-order belief configurations. $\Delta$F1 and relative
gain compare Full ToM with the corresponding lower-order variant.
ECE is global; Brier score is reported on warranted contexts.}
\label{tab:policy-tom-ablation}
\vspace{-1em}
\end{table}

Full ToM-grounded friction improves intervention F1 under both objectives,
from $0.368$ to $0.416$ for FAR and from $0.400$ to $0.417$ for FTR,
corresponding to relative gains of $13.0\%$ and $4.3\%$. Because the
lower-order variants retain the same objectives, data, and surface and
first-order channels, this provides direct policy-level evidence that
second-order belief structure contributes beyond generic friction
conditioning.

The calibration effects are mixed. Full ToM improves FAR's global ECE,
whereas lower-order FTR achieves lower global ECE, and both lower-order
variants obtain slightly lower warranted-context Brier scores. We therefore
interpret the second-order contribution as an improvement in intervention
behavior rather than a general calibration advantage. This complements the
signal-level ablation in \S\ref{sec:validation:channels}, where removing the
second-order channel reduces misunderstanding recall from approximately
$65\%$ to $26\%$.
\paragraph{Robustness to friction-module constants.}
The findings are not tied to a single set of hand-specified constants.
Uniform $F^{-}$ weighting preserves the relevant configuration ordering,
while the proposed weights sharpen cross-class separation by $11.9\%$;
perturbing component weights and scalar mappings by up to $\pm20\%$ does
not reverse the key ordering or downstream conclusions. Perturbations of
the $F^{+}$ utility table likewise do not reverse the inter-method ranking.
We therefore treat these values as structurally motivated surrogates rather
than fitted optima; full sensitivity results appear in
Appendix~\ref{app:sensitivity}.

These staged controls identify three distinct effects. Friction-informed pair
construction improves over friction-blind training; direct per-instance
conditioning provides a further gain; and, holding the objective fixed,
second-order ToM improves intervention F1 beyond lower-order friction alone.
The mixed calibration results also show that the benefit of ToM is
metric- and objective-dependent rather than uniform.

\vspace{-.5em}

\subsection{Training, Prompting, and Test-Time Belief Access}
\vspace{-.5em}
\label{sec:prompting-access}
We compare friction-conditioned training with the untrained backbone and
inference-time prompting. Under the standard inference condition, each input
includes an \emph{observable belief-state preamble}: a compact,
schema-derived summary of the epistemic information available to the
participant-agent, appended to the dialogue history. As a prompt-engineering
baseline, we also evaluate in-context learning (ICL) with the untrained
Qwen3.5-27B~\cite{qwen3.5}: five worked examples (two divergent and three aligned) plus an
instruction to ask for intervention, with no
fine-tuning. This tests whether the intervention behavior learned through FPO
can instead be elicited at inference time. Table~\ref{tab:prompting-access}
compares these conditions and reports the effect of removing the belief
preamble from each trained checkpoint.

\begin{table}[t]
\centering
\small
\setlength{\tabcolsep}{4pt}
\begin{tabular}{lccc}
\toprule
Condition & F1 & ECE & $\Delta$F1 masked \\
\midrule
Base + ToM preamble & $0.410$ & $\mathbf{0.146}$ & $-0.060$ \\
DPO & $0.344$ & $0.174$ & $-0.014$ \\
FAR & $0.416$ & $0.152$ & $-0.073$ \\
FTR & $0.417$ & $0.189$ & $-0.068$ \\
\midrule
ICL (random exemplars) & $\mathbf{0.510}$ & $0.175$ & --- \\
ICL (friction scaffold) & $0.490$ & $\mathbf{0.146}$ & --- \\
\bottomrule
\end{tabular}
\caption{Zero-shot, trained, and in-context conditions on the held-out
test set. F1 and ECE use the intact belief preamble; $\Delta$F1 reports
the change when that preamble is removed from the same checkpoint.
The ICL conditions use a $2$--$3$k-token scaffold and are not included
in the masking comparison.}
\label{tab:prompting-access}
\vspace{-1.5em}
\end{table}

\paragraph{Base-policy competence and in-context learning.}
With the observable belief preamble, the untrained backbone reaches F1
$0.410$, comparable to FAR and FTR under this metric and above DPO. Thus,
substantial intervention competence is already accessible to the base model
when the relevant epistemic state is made explicit. ICL raises F1 further to
$0.490$--$0.510$, but random exemplars perform comparably to the
friction-informed scaffold. The ICL gain therefore appears to reflect generic
exemplar priming rather than a friction-specific prompting advantage. These
results suggest that FAR and FTR primarily preserve and condition existing
intervention competence during adaptation rather than creating it from
scratch.

\paragraph{Belief-preamble masking.}
Removing the observable preamble reduces F1 by $0.060$ for the base model
and by $0.073$ and $0.068$ for FAR and FTR, respectively. The comparable
drops show that these policies continue to rely on explicit epistemic
information at inference rather than fully internalizing it during training.
Under masking, FAR and FTR nevertheless remain slightly above DPO
($0.343$/$0.349$ vs.\ $0.330$), indicating some residual robustness when the
preamble is unavailable. DPO changes little under masking, consistent with
its weaker dependence on instance-specific epistemic information.

These controls refine the role of ToM-grounded friction:
friction-conditioned training preserves context-sensitive intervention
behavior without requiring a long ICL scaffold, but its strongest performance
still depends on the compact observable epistemic representation at test
time. Full prompting specifications and masked-condition results appear in
Appendix~\ref{app:prompting-access}.
\vspace{-.5em}
\section{Conclusion}
\vspace{-.5em}

Interactive systems are commonly optimized for response quality, yet successful collaboration also depends on whether interlocutors maintain sufficiently compatible interpretations throughout an interaction. Our results show that this distinction is consequential not only for evaluation, but also for representation and learning. By tracking, for each referring expression, what each participant intends, understands, and believes about the other's interpretation, the four-identifier ToM schema renders otherwise latent misalignment computationally accessible. Within our extraction framework, removing the second-order component substantially reduces misunderstanding recovery, indicating that lower-order information alone does not capture the relational structure needed to identify many dialogue breakdowns.

Once represented explicitly, these asymmetries can serve as supervisory signals rather than merely diagnostic annotations. FAR and FTR incorporate ToM-grounded friction through distinct optimization mechanisms, yet both improve intervention behavior and warranted-context calibration relative to DPO while substantially increasing training stability. The lower-order ablations further indicate that these gains are not attributable to generic friction conditioning alone, but depend in part on second-order information about participants' interpretations of one another. 

More broadly, these findings motivate a view of dialogue alignment that extends beyond preference optimization over utterances toward the regulation of evolving belief relations. From this perspective, intervention is particularly valuable when locally coherent interaction masks incompatible interpretations that would otherwise persist. Because $F^-$ is defined over epistemic state rather than tied to a specific optimization objective, future work can investigate its use as a reweighting, reward-shaping, or control signal beyond the FPO variants considered here. Ultimately, successful dialogue alignment requires interactive systems to recognize and repair emerging misalignment before apparent coordination consolidates into shared error.







\section*{Limitations}Our experiments are conducted on the HCRC MapTask corpus, a controlled navigation task chosen because its dense referential grounding and mismatched participant maps provide an ideal testbed for evaluating latent epistemic alignment. While this structured environment allows for precise evaluation of the four-identifier ToM schema, exploring its generalization to less constrained open-domain or multi-party dialogue remains an objective for future work. Additionally, the misunderstanding rate in MapTask (5.5--9.1\%) reflects its specific task asymmetric design and may vary in other collaborative corpora.

The corpus provides first-order perspectivist annotations but no gold
second-order mental-state labels. We therefore infer the latter with GPT-5.
A blind human pilot shows substantial agreement with these predictions
($\kappa=0.80/0.77$), supporting their use as supervisory signals, but it
does not establish them as ground-truth mental states. More generally, our
evidence for genuinely non-merged perspective-taking remains suggestive
rather than conclusive. The signal and policy ablations show that the
second-order representation contributes beyond the lower-order channels
tested here, but they cannot rule out reliance on correlated linguistic or
interactional cues rather than genuine mentalizing.

Friction is computed offline from reconstructed epistemic trajectories rather
than estimated incrementally during live interaction. Full cross-party ToM
information is used to construct this supervisory friction signal, but it is
not a direct policy input: during both training and standard inference, the
policy receives dialogue history plus the same compact agent-observable belief
projection. The masking experiment shows that performance remains partly
dependent on this representation at inference. Fully online estimation,
error propagation over extended interaction, and rollout-based variants such
as GRFR therefore remain outside the present evaluation.

Finally, the generality of the friction formulation remains untested beyond
the objectives and intervention space studied here. Although $F^-$ is defined
over epistemic state rather than a particular optimization objective, we have
not established that it transfers as a reweighting, reward-shaping, or control
signal to other preference- or reinforcement-learning methods. $F^+$ is more
explicitly tied to the inherited FPO action space
(\texttt{clarify}/\texttt{verify}/\texttt{redirect}/\texttt{refuse}); using
it in other settings would require defining productive friction over the
corresponding action space. The hand-set friction constants are robust to the
perturbations tested here, but this does not establish their optimality.

\section*{Acknowledgments}

This research was supported by the NSF National AI Institute for Student-AI Teaming (iSAT) under grant DRL 2454151. The opinions expressed are those of the authors and do not represent views of the NSF.

We used generative AI tools during the preparation of this work. Specifically, assistance from the Claude, ChatGPT, and Gemma 4 model families was used to support literature search, research-related coding, and language refinement, including tone and style checking and paraphrasing. All literature identified with AI assistance was independently verified and validated by the authors for relevance, factual accuracy, and the correctness of citation content and formatting. The authors reviewed all AI-assisted outputs and take full responsibility for the accuracy and integrity of the submitted work.

\section*{Potential Risks}
The primary artifact introduced in this work is a friction signal derived from Theory-of-Mind inference over dialogue. While the immediate application is alignment training for collaborative dialogue agents, the same underlying machinery could, in principle, be deployed to detect and strategically exploit belief asymmetries between interlocutors rather than to repair them. We explicitly flag this dual-use potential for adversarial manipulation; however, we note that our framework is structurally engineered for epistemic stewardship, optimizing exclusively for transparency, verification, and mutual repair signals.

Additionally, our second-order belief inference pipeline relies on GPT-5, a proprietary model whose latent representations may reflect demographic or stylistic biases present in its training data. Systematic errors in ToM attribution—particularly speaker-side belief inference, where baseline accuracy is naturally bounded —could cascade into miscalibrated friction signals that penalize appropriate conversational progress or incentivize redundant, costly interventions. Consequently, we strongly advise that any downstream adaptation of friction-conditioned policies in high-stakes domains maintain strict human-in-the-loop oversight to audit model calibration during continuous interaction.


\bibliography{custom}

\appendix

\section{Friction Specification and Sensitivity}

\subsection{$F^-$ Component Surrogates}
\label{app:f-}
Table~\ref{tab:friction-surrogates-app} gives the complete scalar mappings
used to instantiate the four components of $F^-$. All values are deterministic
surrogates normalized to $[0,1]$ rather than parameters fitted to the
evaluation data.

\begin{table}[h]
\centering
\small
\setlength{\tabcolsep}{4pt}
\begin{tabular}{@{}llr@{}}
\toprule
Component & Condition & Score \\
\midrule
\multirow{4}{*}{\textsc{Unc}}
 & committed & 0.05 \\
 & hesitant  & 0.55 \\
 & withheld  & 0.70 \\
 & absent    & 0.40 \\
\midrule
\multirow{6}{*}{\textsc{Contr}}
 & contradict       & 0.85 \\
 & repair           & 0.65 \\
 & request\_clarify & 0.55 \\
 & pause            & 0.20 \\
 & acknowledge      & 0.05 \\
 & none             & 0.00 \\
\midrule
\multirow{4}{*}{\textsc{Haz}}
 & existence asymmetry     & 0.65 \\
 & multiplicity divergence & 0.55 \\
 & lexical variant         & 0.30 \\
 & otherwise               & 0.00 \\
\midrule
\multirow{4}{*}{\textsc{ValConf}}
 & no divergence            & 0.00 \\
 & diverged, neither aware  & 0.95 \\
 & diverged, one aware      & 0.55 \\
 & diverged, both aware     & 0.25 \\
\bottomrule
\end{tabular}
\caption{Complete scalar mappings used by the $F^-$ friction components.}
\label{tab:friction-surrogates-app}
\end{table}

\textsc{Unc} is derived from addressee uptake quality.
\textsc{Contr} uses the strongest active repair-oriented surface signal,
rather than averaging multiple markers. \textsc{Haz} uses the strongest
applicable map-level risk condition and is additionally scaled by remaining
task length using $\textit{len\_mult}\in\{0.7,1.0,1.3\}$.
\textsc{ValConf} is determined by the four-identifier belief relations:
referential divergence is
$\textcircled{\small1}\neq\textcircled{\small2}$;
speaker awareness is
$\textcircled{\small3}\neq\textcircled{\small1}$;
and addressee awareness is
$\textcircled{\small4}\neq\textcircled{\small2}$ or is surfaced through
repair-oriented signals.



\subsection{Sensitivity to Design-Choice Constants}
\label{app:sensitivity}
The friction module contains two sets of hand-specified constants:
the component weights
$w$=(0.15,0.20,0.30,0.35) and the categorical-to-scalar mappings
described above. We perturb individual and joint component weights by
$\delta\in\{-20\%,-10\%,+10\%,+20\%\}$ and separately apply uniform
perturbations to the scalar mappings. For each condition, we recompute
$F^-$ over all 278 held-out instances, the configuration ordering, and
the FTR coefficient
$\beta(h)=1/(\epsilon_0+\kappa F^-(h))$.

The perturbations preserve the structurally informative configuration
ordering and maintain a substantial gap between
\texttt{silent\_intent\_fixing} and \texttt{aligned\_deep}
(Table~\ref{tab:sensitivity}). The FTR risk contrast likewise remains
stable, while perturbing the $F^+$ utility values changes InfoEff
quantitatively without reversing the inter-method ordering.

\begin{table*}[h!] \centering \small \setlength{\tabcolsep}{6pt} \begin{tabular}{lcccc} \toprule Perturbation & Cfg Kendall $\tau$ & Cfg gap & $\beta$-ratio & InfoEff rank \\ \midrule Reference & $0.867^\star$ & $+0.46$ & $2.12$ & DPO $<$ FAR/FTR \\ $\delta{=}-20\%$ (uniform $W$) & $0.867^\star$ & $+0.37$ & $1.93$ & preserved \\ $\delta{=}-10\%$ (uniform $W$) & $0.867^\star$ & $+0.42$ & $2.03$ & preserved \\ $\delta{=}+10\%$ (uniform $W$) & $0.867^\star$ & $+0.51$ & $2.17$ & preserved \\ $\delta{=}+20\%$ (uniform $W$) & $0.867^\star$ & $+0.56$ & $2.18$ & preserved \\ $\delta{=}\pm 20\%$ (single $w_k$, worst case) & $\geq 0.733$ & $+0.40$ & $1.93$ & preserved \\ $\delta{=}\pm 20\%$ (scalar map, uniform) & $\geq 0.733$ & $+0.37$ & $1.93$ & preserved \\ \bottomrule \end{tabular} \caption{Sensitivity to friction-module constants ($n{=}278$). \textbf{Cfg $\tau$}: Kendall rank-correlation of per-configuration mean $F^-$ vs.\ reference. \textbf{Cfg gap}: $F^-(\texttt{silent\_intent\_fixing}) - F^-(\texttt{aligned\_deep})$. \textbf{$\beta$-ratio}: $\beta(\texttt{aligned\_deep})/\beta(\texttt{silent\_intent\_fixing})$ under FTR. \textbf{InfoEff rank}: whether DPO $<$ \{FAR,FTR\} survives. $^\star$Single near-zero tie between \texttt{multiplicity\_suspicious} and \texttt{aligned\_with\_repair} ($\Delta{=}0.005$); all five structurally informative pairs are concordant in every perturbation.} \label{tab:sensitivity} \end{table*}

We additionally compare the proposed weighting with the non-informative
uniform alternative $w=(0.25,0.25,0.25,0.25)$. Both preserve the
pending--misunderstood ordering
($\Delta=+0.218$ proposed; $+0.179$ uniform), while the proposed weighting
increases cross-class separation by $11.9\%$ ($0.463$ vs.\ $0.414$).
We therefore treat the constants as robust structural surrogates rather
than claiming that they are optimal.

\subsection{$F^+$ Productive-Friction Utility}
\label{app:f+}

In the offline setting, $\textsc{InfoGain}(h,a)$ is implemented as the lookup table in Table~\ref{tab:f-plus-table}. Diagnostic actions (\texttt{clarify}, \texttt{verify}) receive the highest values on divergent states, especially \texttt{silent\_intent\_fixing}, where neither participant locally detects the mismatch. Corrective actions (\texttt{redirect}, \texttt{refuse}) are scored lower because they assume the source of divergence is already identifiable and impose greater conversational cost. Aligned states receive near-zero values, encoding that unnecessary intervention yields little epistemic gain.

\begin{table}
[t] \centering \small \setlength{\tabcolsep}{4pt} 
\resizebox{\columnwidth}{!}{%
\begin{tabular}{lrrrr} \toprule Configuration & clar. & verif. & redir. & refuse \\ \midrule \multicolumn{5}{@{}l}{\textit{Divergent:}} \\ \texttt{silent\_intent\_fixing} & $0.85$ & $\mathbf{0.95}$ & $0.55$ & $0.20$ \\ \texttt{asym\_aware\_speaker\_only} & $0.85$ & $0.80$ & $0.45$ & $0.15$ \\ \texttt{asym\_aware\_addressee\_only} & $0.85$ & $0.70$ & $0.45$ & $0.15$ \\ \texttt{open\_dispute} & $0.70$ & $0.80$ & $0.50$ & $0.10$ \\ \texttt{multiplicity\_suspicious} & $0.75$ & $\mathbf{0.85}$ & $0.40$ & $0.10$ \\ \midrule \multicolumn{5}{@{}l}{\textit{Aligned:}} \\ \texttt{aligned\_deep} & $0.05$ & $0.10$ & $0.05$ & $0.02$ \\ \texttt{aligned\_with\_repair} & $0.40$ & $0.40$ & $0.20$ & $0.05$ \\ \midrule \multicolumn{5}{@{}l}{\textit{Underdetermined:}} \\ \texttt{underspecified} & $\mathbf{0.90}$ & $0.50$ & $0.25$ & $0.05$ \\ \texttt{existence\_query} & $0.20$ & $0.30$ & $0.10$ & $0.05$ \\ \bottomrule \end{tabular}} \caption{$F^+(h,a)$ values by referential configuration and intervention type.} \label{tab:f-plus-table} \vspace{-1em} 
\end{table}

\section{ToM Extraction and Validation}
\label{app:tom-extraction}

\subsection{Preliminary Representation Alternatives}

Two preliminary formulations motivated the final four-identifier schema.
First, local-context belief prediction performed below a constant-True
baseline. In particular, latent divergence was frequently not identifiable
from the immediate referring-expression context and became attributable only
after subsequent repair, contradiction, or task failure.

We therefore adapted the retrospective grounding cascade of
\citet{li2026groundedmisunderstandingsasymmetricdialogue}, allowing later
dialogue evidence to inform the belief state attributed at the original
reference. Although this improved contextual access, the representation
remained structurally insufficient: grounding was encoded as an intra-party
binary outcome, whereas silent divergence occurs when two participants
confidently ground the same expression to different referents.

These preliminary results motivated two properties of the final
representation: beliefs are attributed to the state at the time of the
reference while allowing retrospective evidence for offline reconstruction,
and first- and second-order beliefs are represented separately across
participants.

\subsection{Human Annotation Protocol}

\label{app:human-annotation}

Because the corpus contains no native second-order belief labels, we
independently evaluated GPT-5's second-order predictions in a blind human
annotation pilot. We conducted a blind human-annotation pilot on a stratified sample of $60$ referring expressions drawn from the held-out test partition ($n_{\text{test}}=761$). Sampling enforced a minimum of eight items per configuration class whenever class frequency permitted, ensuring coverage of all eight epistemic configurations, including divergent cases such as \texttt{silent\_intent\_fixing} and \texttt{multiplicity\_suspicious}. The annotator is an author of this paper, fluent in English and familiar with both the HCRC MapTask corpus and the four-identifier ToM schema introduced in \S\ref{sec:tom-module}. The pilot was designed to evaluate whether GPT-5 second-order predictions aligned with human-interpretable referential judgments under the proposed schema, rather than to establish inter-annotator agreement for the schema itself.

For each item, the annotator had access to the dialogue context preceding the target referring expression, the target expression itself, speaker role (giver or follower), dialogue and map identifiers, and the corresponding MapTask map images for both the giver and the follower. This matches the GPT-5 prompting condition (\S\ref{sec:tom-module}) on the dialogue-text and map-vision channels; the only information unavailable to the annotator was the inline dialogue-act annotation stream additionally consumed by GPT-5 for surface-signal grounding. GPT-5 outputs were withheld throughout annotation to prevent anchoring effects. The worksheet header provided the following written instructions:

\begin{quote}\small \emph{For each row, check the corresponding map and read the} \texttt{dialogue\_context} \emph{and} \texttt{target\_re\_text}. \emph{Second-order ToM refers to what each participant believes the other participant meant or understood. Fill in predicted landmark identifiers for} \texttt{user\_speaker\_tom\_landmark} \emph{and} \texttt{user\_addressee\_tom\_landmark}. \emph{Use corpus landmark identifiers where possible (e.g.,} \texttt{m3\_sandstone\_cliffs\#0@g}\emph{); free-text descriptions are also acceptable. Enter} \texttt{unknown} \emph{if the referent cannot be determined from dialogue evidence alone.} \end{quote} No additional verbal instructions were provided. Annotation was completed in a single session of approximately two hours, producing $120$ second-order labels ($60 \times 2$ identifier slots). The worksheet was edited in spreadsheet format and exported to CSV.

Agreement was computed using an automated scoring script that canonicalized map prefixes (\texttt{m3\_}, \texttt{m10\_}), instance indices (\texttt{\#0}), map-role suffixes (\texttt{@f}, \texttt{@g}), common spelling variants (\texttt{wheat\_fields}/\texttt{wheatfields}), and leading articles before exact-match comparison. Cohen's $\kappa$ is reported over the canonicalized labels. The HCRC MapTask corpus \citep{anderson1991hcrc} is publicly released for academic research and contains no personally identifying or sensitive material. The annotation task involved only referential interpretation over existing corpus data. The annotator received no separate compensation and performed the work as part of routine research activity. Because the pilot involved only author annotation over publicly available data, IRB review was not sought.

\section{Data Preprocessing and Split Construction}
\label{app:data-preprocessing}

For each MapTask dialogue, inline referring-expression annotations are
aligned with the corresponding giver- and follower-specific belief
annotations. Referring expressions spanning multiple same-speaker
continuation turns are stitched into a single target span while intervening
utterances are preserved, so reconstruction does not remove conversational
evidence occurring within the reference. Each resulting instance retains the
dialogue history through the target expression and the following eight
utterances used for completion-side evaluation.

Dialogues are partitioned at the dialogue level rather than the
referring-expression level. Splitting is stratified by misunderstood-RE
density to preserve a range of dialogue difficulty, with development
dialogues selected near the distribution median to reduce prompt selection
from being dominated by unusually easy or difficult interactions.

The representation-level split contains 2,705 training REs and 761 held-out
test REs. Before policy adaptation, we remove instances that are
underspecified for the required referential comparison and explicit
existence-query cases, for which questioning itself reveals the relevant
information asymmetry. The resulting policy cache contains 663 training
instances and 278 held-out evaluation instances. Thus, the larger split is
used for extraction and friction-signal validation, whereas the filtered
subset defines the policy-training and downstream-evaluation setting.


\section{Offline Trainer Implementation}
\label{app:training}

\subsection{FAR Offline Adaptation}
\label{app:far}
Applying FAR to the offline corpus requires two adaptations. First, on
divergent rows the observed corpus response may itself be a silent
acknowledgment. Assigning a negative task reward to this completion without
an alternative target caused training collapse in pilot runs. We therefore
retarget divergent rows to the corresponding corpus-retrieved repair
completion and use $R_{\text{task}}=+1$ after retargeting.

Second, the original policy-gradient formulation assumes on-policy rollouts,
which are unavailable in the fixed RE-level corpus. We therefore use
REINFORCE with running-mean baseline subtraction
\citep{williams1992simple} and reward clipping.

\subsection{FPP Offline Adaptation}
\label{app:fpp}
Three adjustments are required for FPP in the offline setting. First,
$F^+(\texttt{answer})$ is not defined by the intervention lookup table.
For aligned pairs we therefore use
\[
F^+(\texttt{answer}\mid\text{aligned})
 = 1-\max_a F^+(h,a),
\]
without modifying the underlying friction module.

Second, because $F^-$ varies substantially across epistemic configurations,
we weight each pair by
\[
w(h)\propto F^-(h)+\varepsilon,
\]
so that higher-risk rows contribute more strongly to the preference
objective.

Third, only $9.3\%$ of divergent corpus repair templates are recognized by
the rule-based intervention classifier without lexical augmentation.
For divergent chosen completions that do not already expose the relevant
class, we prepend a minimal marker:
\textit{sorry}$\rightarrow$\texttt{clarify},
\textit{so}$\rightarrow$\texttt{verify},
\textit{no}$\rightarrow$\texttt{redirect}, and
\textit{i can't}$\rightarrow$\texttt{refuse}.
\subsection{FTR Implementation}
\label{app:ftr}
For the single-completion offline setting, exact sequence-level KL is not
directly available. We use the non-negative $k_3$ estimator
\[
\widehat{\mathrm{KL}} = r-1-\log r,
\]
where $r$ is the policy/reference likelihood ratio. Pilot experiments with
signed and squared log-ratio estimators did not preserve the desired
asymmetry with respect to decreases in gold-completion probability.

We set $\mathrm{Risk}(h)=F^-(h)$ directly rather than fitting a separate
risk predictor. With $\epsilon_0=1.0$ and $\kappa=3.0$,
\[
\beta(h)=\frac{1}{\epsilon_0+\kappa F^-(h)}
\]
falls in approximately $[0.36,1.0]$ over the observed friction range.
Tighter regimes over-anchored the policy in pilot runs and increased
calibration error.


\section{Additional Evaluation}

\subsection{Per-Seed Stability}
\label{app:multiseed}

Repeating the full train-and-evaluate protocol from three independent QLoRA
initializations (seeds $42$, $123$, and $7$) yields two complementary
patterns. First, the FAR/FTR advantage replicates at the individual-run
level: all six same-seed comparisons against DPO are significant under
McNemar tests ($p\leq3\times10^{-3}$), with net margins of
$+14.4$--$+42.5$ percentage points for FAR and
$+13.3$--$+36.7$ for FTR.

Second, DPO is substantially more variable across initializations. Its F1
ranges from $0.152$ to $0.398$ (std.\ $0.106$), and warranted-context ECE
reaches $0.646$ in the weakest run, whereas FAR and FTR have F1 standard
deviations of $0.023$ and $0.009$. Because only three seeds are available,
aggregate seed-paired tests are underpowered; we therefore interpret the
cross-run result primarily as evidence of greater training stability rather
than as a precise estimate of between-method variance.

\begin{table}[t]
\centering
\small
\setlength{\tabcolsep}{3pt}
\begin{tabular}{llcccc}
\toprule
Method & Seed & F1 & ECE & ECE\textsubscript{warr} & Net win vs DPO \\
\midrule
\multirow{3}{*}{DPO}
 & 42  & $0.344$ & $0.174$ & $0.432$ & --- \\
 & 123 & $0.152$ & $0.233$ & $0.646$ & --- \\
 & 7   & $0.398$ & $0.299$ & $0.491$ & --- \\
\midrule
\multirow{3}{*}{FAR}
 & 42  & $0.439$ & $0.147$ & $0.234$ & $+14.4$ \\
 & 123 & $0.409$ & $0.142$ & $0.174$ & $+24.5$ \\
 & 7 & $0.383$ & $0.178$ & $0.265$ & $+42.5$ \\
\midrule
\multirow{3}{*}{FPP}
 & 42  & $0.000$ & $0.161$ & $0.512$ & $-38.8$ \\
 & 123 & $0.000$ & $0.161$ & $0.512$ & $-21.6$ \\
 & 7   & $0.000$ & $0.161$ & $0.512$ & n/a \\
\midrule
\multirow{3}{*}{FTR}
 & 42  & $0.417$ & $0.189$ & $0.178$ & $+13.3$ \\
 & 123 & $0.434$ & $0.230$ & $0.205$ & $+17.6$ \\
 & 7   & $0.438$ & $0.188$ & $0.193$ & $+36.7$ \\
\bottomrule
\end{tabular}
\caption{Per-seed breakdown of Table~\ref{tab:multiseed}. FPP's F1/ECE/ECE\textsubscript{warr} are identical across seeds by construction; its win rate vs.\ DPO still varies because DPO's completions change per seed.}
\label{tab:multiseed-perseed}
\end{table}

\subsection{Prompting and Preamble Masking}
\label{app:prompting-access}

\paragraph{Belief-state preamble.} Every \texttt{prompt\_user} field is built from a fixed template: a first-order belief block (the speaker's own anchoring of the current referring expression to a landmark instance on their map, derived from the schema-extracted \texttt{speaker\_belief}/\texttt{addressee\_belief} identifiers) followed by a \texttt{Dialogue history:} marker and the dialogue so far. No second-order (\texttt{speaker\_tom}/\texttt{addressee\_tom}) content appears in the prompt at inference under any trained policy; those channels are used only to construct the friction-conditioned training objective. An example (test set, configuration \texttt{multiplicity\_suspicious}):

\begin{quote}
\small\ttfamily
Your current belief state:\\
\ -\ You have anchored the partner's reference to the wheat fields (instance \#1) on your map (upper).\\
\ -\ You believe the partner meant the wheat fields (instance \#1) on your map (upper).\\[4pt]
Dialogue history:\\
G: right so the start is at the top left-hand side of the page\\
F: uh-huh\\
{[}\ldots{]}\\
G: no hold on wait a minute am i right under the sandstone cliffs
\end{quote}

The masked condition removes everything before the \texttt{Dialogue history:} marker verbatim, leaving the dialogue turns (and all other prompt content) untouched.

\paragraph{In-context learning scaffold.} The ICL condition prompts the untrained base policy with a fixed instruction plus $N{=}5$ few-shot examples (seed $42$; $2$ drawn from divergent configurations \{\texttt{silent\_intent\_fixing}, \texttt{multiplicity\_suspicious}, \texttt{asymmetric\_aware\_*}, \texttt{open\_dispute}\}, $3$ from aligned configurations, sampled from the training split), each example truncated to its dialogue and gold completion:

\begin{quote}
\small\ttfamily
Below are examples of correct responses in a MapTask collaboration. Study them, then respond in the same style.\\[4pt]
Example 1:\\
Dialogue:\\
{[}dialogue excerpt{]}\\
Correct response: {[}gold completion{]}\\[4pt]
{[}\ldots \ 4 more examples\ldots{]}\\[4pt]
Now respond to the following dialogue. If the speaker's reference is ambiguous or you cannot resolve it from the dialogue evidence and your map, ask a brief clarifying question. Otherwise acknowledge and proceed.
\end{quote}

This scaffold ($\sim$2--3k tokens) is prepended to each test row's own belief preamble and dialogue. The \emph{generic} control (ICL random exemplars) uses the same five-slot template with exemplars sampled irrespective of configuration and the final instruction sentence removed, isolating the contribution of exemplar priming from the friction-specific instruction.
\end{document}